\documentclass[11pt,a4paper]{article}
\usepackage[hyperref]{acl2018}
\usepackage{times}
\usepackage{latexsym}
\usepackage{subcaption}

\usepackage{url}

\aclfinalcopy % Uncomment this line for the final submission
\def\aclpaperid{1627} %  Enter the acl Paper ID here

\usepackage{graphicx}
\usepackage{amsmath, amssymb}
\usepackage{float}
\usepackage{array, booktabs}
\usepackage{comment}

\usepackage{bm}
\usepackage{mathrsfs}

\DeclareMathOperator*{\argmax}{arg\,max}

\usepackage{CJKutf8}

\title{Neural Adversarial Training for Semi-supervised \\
 Japanese Predicate-argument Structure Analysis}

\author{\hspace{4em}Shuhei Kurita\hspace{-0.1mm}$^\$$\hspace{-0.1mm}$^\dagger$\hspace{-0.1mm}$^{\ddagger}$
       \\\And Daisuke Kawahara\hspace{-0.1mm}$^\$$\hspace{-0.1mm}$^\dagger$\\
         $^\$$Graduate School of Informatics, Kyoto University \\
     $^\dagger$CREST, JST~~~~$^{\ddagger}$ACT-I, JST\\
         {\tt \{kurita, dk, kuro\}@nlp.ist.i.kyoto-u.ac.jp}
       \\\And \hspace{-4em}Sadao Kurohashi\hspace{-0.1mm}$^\$$\hspace{-0.1mm}$^\dagger$\hspace{-0.1mm}$^{\ddagger}$\\
}

\date{}
\begin{document}
\maketitle
\begin{abstract}
Japanese predicate-argument structure (PAS) analysis involves zero
anaphora resolution, which is notoriously difficult. To improve the
performance of Japanese PAS analysis, it is straightforward to increase
the size of corpora annotated with PAS. However, since it is
prohibitively expensive, it is promising to take advantage of a large
amount of raw corpora. In this paper, we propose a novel Japanese PAS
analysis model based on semi-supervised adversarial training with a raw
corpus. In our experiments, our model outperforms existing
state-of-the-art models for Japanese PAS analysis.

%Annotated corpus is very limited.
%Language have its own structure, such as grammars or selectional preference of predicates and arguments,
%and the model could learn the hidden structures from unannotated corpus.
%We propose adversarial learning and discriminator for learning selectional preference.
%our model successfully capture the SP and achieve SOTA.
\end{abstract}

% ロジック
% pro-drop languageの解析の難しさ
% Japanese PAS
%

\begin{table*}[t]
%\begin{center}
\hspace{-0.7em}\includegraphics[scale=0.87,clip]{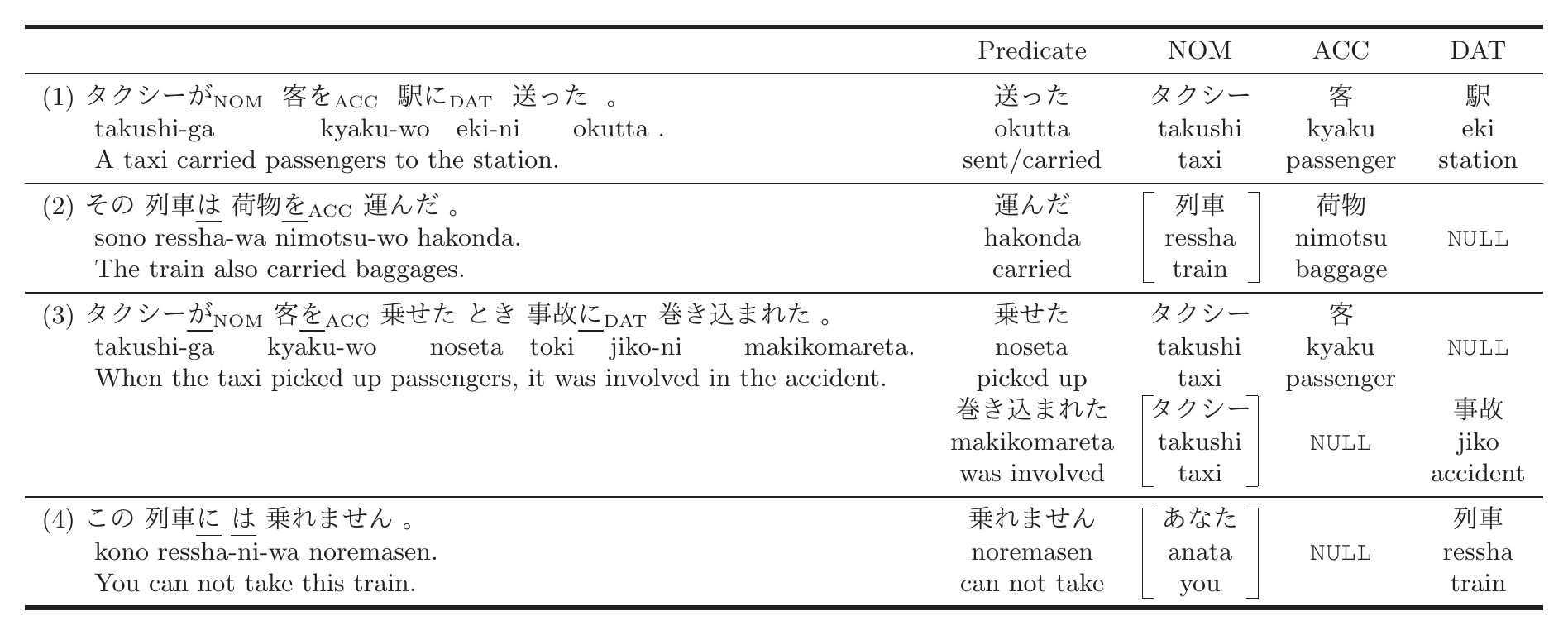}
\vspace{-2em}
    \caption{
Examples of Japanese sentences and their PAS analysis.
In sentence (1), case markers (
{\small \begin{CJK*}{UTF8}{ipxm}\underline{が}\end{CJK*}(ga)},
{\small \begin{CJK*}{UTF8}{ipxm}\underline{を}\end{CJK*}(wo)}, and
{\small \begin{CJK*}{UTF8}{ipxm}\underline{に}\end{CJK*}(ni)}
) correspond to NOM, ACC, and DAT.
In example (2),
the correct case marker is hidden by the topic marker
{\small \begin{CJK*}{UTF8}{ipxm}\underline{は}\end{CJK*} (wa)}.
In sentence (3), the NOM argument of the second predicate 
{\small \begin{CJK*}{UTF8}{ipxm}巻き込まれた\end{CJK*} (was involved)},
is dropped.
{\usefont{T1}{pcr}{m}{n} NULL}
indicates that the predicate does not have the corresponding case argument or that the case argument is not written in the sentence.
    }
    \label{table:examples}
%\end{center}
\end{table*}

\section{Introduction}

        %%% (3) \begin{CJK*}{UTF8}{ipxm}客を~乗せた~タクシーも~事故に~巻き込まれた~。\end{CJK*} & \begin{CJK*}{UTF8}{ipxm}乗せた\end{CJK*} &\begin{CJK*}{UTF8}{ipxm}タクシー\end{CJK*} & \begin{CJK*}{UTF8}{ipxm}客\end{CJK*} & {\usefont{T1}{pcr}{m}{n} NULL}  \\
        %%% A taxi carrying passengers was also involved in the accident. & carrying & taxi & passenger  & \\
        %%% & \begin{CJK*}{UTF8}{ipxm}巻き込まれた\end{CJK*} &\begin{CJK*}{UTF8}{ipxm}タクシー\end{CJK*} & {\usefont{T1}{pcr}{m}{n} NULL} & \begin{CJK*}{UTF8}{ipxm}事故\end{CJK*} \\
        %%% & was involved & taxi & & accident \\

In pro-drop languages, such as Japanese and Chinese, pronouns are
frequently omitted when they are inferable from their contexts and
background knowledge. The natural language processing (NLP) task for
detecting such omitted pronouns and searching for their antecedents is
called zero anaphora resolution. This task is essential for downstream
NLP tasks, such as information extraction and summarization.

For Japanese, zero anaphora resolution is usually conducted within
predicate-argument structure (PAS) analysis as a task of finding an
omitted argument for a predicate. PAS analysis is a task to find an
argument for each case of a predicate. For Japanese PAS analysis, the \textit{ga}
(nominative, NOM), \textit{wo} (accusative, ACC) and \textit{ni}
(dative, DAT) cases are generally handled.
% These cases are roughly correspond to \textit{who}, \textit{what} and \textit{whom}, respectively.
To develop models for Japanese PAS analysis, supervised learning methods
using annotated corpora have been applied on the basis of
morpho-syntactic clues. However, omitted pronouns have few clues and
thus these models try to learn relations between a
predicate and its (omitted) argument from the annotated corpora. The annotated
corpora consist of several tens of thousands sentences, and it is
difficult to learn predicate-argument relations or selectional preferences from such small-scale
corpora. The state-of-the-art models for Japanese PAS analysis achieve an accuracy
of around 50\% for zero pronouns \cite{ouchi2015,shibata2016,iida2016,ouchi2017,matsubayashi2017}.

A promising way to solve this data scarcity problem is enhancing models
with a large amount of raw corpora.
There are two major approaches to using raw corpora:
extracting knowledge from raw corpora beforehand \cite{sasano2011,shibata2016}
and using raw corpora for data augmentation \cite{tingliu2017}.

In traditional studies on Japanese PAS analysis,
selectional preferences are extracted from raw corpora beforehand and are used in PAS analysis models.
% Selectional preferences are basic knowledges of predicates and their possible arguments.
% For example, in a sentence of English "\textit{Someone draws on pictures on $\phi$.}",
% \textit{$\phi$} is likely paper, canvas, screen, and so on.
% Case frame by \newcite{kawahara2006} is a large and sophisticated collections of selectional preferences extracted from an extremely large web corpus by unsupervised clustering method.
% In case frame, each predicate is classified into several frame with possible candidate arguments.
% %However, the obtained sectional preferences knowledge can not be adapted to Japanese Anaphora resolution tasks
For example, \newcite{sasano2011} propose a supervised model for Japanese
PAS analysis based on case frames, which are automatically acquired from
a raw corpus by clustering predicate-argument structures.
%[\begin{CJK*}{UTF8}{ipxm} case frameの類似研究も\end{CJK*}].
%One problem of case frame is that the obtained 
However, case frames are not based on distributed representations of words and have a data sparseness problem even if a large raw corpus is employed.
%Some of recent work employ neural network based approach
Some recent approaches to Japanese PAS analysis combines neural network models with knowledge extraction from raw corpora.
% \newcite{shibata2016} use a neural network-based approach on selectional preferences and Japanese PAS analysis.
\newcite{shibata2016} extract selectional preferences by an unsupervised method that is similar to negative sampling \cite{mikolov2013}.
They then use the pre-extracted selectional preferences as one of the features to their PAS analysis model.
The PAS analysis model is trained by a supervised method and the selectional preference representations are fixed during training.
%Therefore they don't employ unsupervised or semi-supervised training directly on their PAS analysis neural network model.
Using pre-trained external knowledge in the form of word embeddings has also been ubiquitous.
However, such external knowledge is overwritten in the task-specific training.
% [\begin{CJK*}{UTF8}{ipxm} 外付けかつ間接的で効果が薄まってダメだと言う\end{CJK*}].
% We make use of external knowledge extracted from a raw corpus at the same time with the task-specific training.

The other approach to using raw corpora for PAS analysis is data augmentation.
%Data augmentation is also applied for making use of raw corpus.
\newcite{tingliu2017} generate pseudo training data from a raw corpus and use them for their zero pronoun resolution model.
They generate the pseudo training data by dropping certain words or pronouns in a raw corpus and assuming them as correct antecedents.
% ?? They also extracted query and document for this pseudo training data.
After generating the pseudo training data, they rely on ordinary supervised training based on neural networks.
% [\begin{CJK*}{UTF8}{ipxm} noisy?でダメだと言う\end{CJK*}].

%\newcite{ouchi2017} doesn't employ selectional preferences model and do not make use of 

In this paper, we propose a neural semi-supervised model for Japanese PAS analysis.
We adopt neural adversarial training to directly exploit the advantage of using a raw corpus.
Our model consists of two neural network models: a generator model of Japanese PAS analysis and a so-called ``validator'' model of the generator prediction.
The generator neural network is a model that predicts probabilities of candidate arguments of each predicate using RNN-based features and a head-selection model \cite{zhang-cheng-lapata2017}.
The validator neural network gets inputs from the generator and scores them.
This validator can score the generator prediction even when PAS gold labels are not available.
We apply supervised learning to the generator and unsupervised learning to the entire network using a raw corpus.
% We train these two neural network models on raw corpus with a semi-supervised method, our model can learn a large number of selectional preferences.

Our contributions are summarized as follows:
(1) a novel adversarial training model for PAS analysis;
(2) learning from a raw corpus as a source of external knowledge; and
% (3) neural head selection and RNN-based PAS analysis model;
(3) as a result, we achieve state-of-the-art performance on Japanese PAS analysis.

\begin{comment}

On the other hand, Japanese anaphora resolution strongly rely on the knowledge of predicates and arguments preferences.
%Predicate argument structure analysis(PAS)
This knowledge of predicates and arguments is known as selectional preferences
[\begin{CJK*}{UTF8}{ipxm}shibata2016を引用してもいいが、最初にselectional preferencesと言い始めた論文がほしい\end{CJK*}].
For example,
%\item ``\begin{CJK*}{UTF8}{ipxm}当店で 焼く\end{CJK*}''
%\begin{itemize}
%\item bake \textit{what} by a oven
%\end{itemize}
%\item ``\begin{CJK*}{UTF8}{ipxm}手を ($\phi$に) 焼く\end{CJK*}''
%\begin{itemize}
%\item having trouble with \textit{whom}
%\end{itemize}
\begin{itemize}
\item ``\begin{CJK*}{UTF8}{ipxm}オーブンで~($\phi$を)~焼く\end{CJK*}''
\begin{itemize}
\item bake (\textit{what}) in oven
\end{itemize}
\item ``\begin{CJK*}{UTF8}{ipxm}絵を~($\phi$に)~描く\end{CJK*}''
\begin{itemize}
\item drawing a picture (on \textit{what})
\end{itemize}
\end{itemize}
In the first Japanese phrase, what is baked in oven is dropped.
However we can limit the candidate arguments that are related to dishes, such as cookies or breads.
In the second example, we can infer \begin{CJK*}{UTF8}{ipxm}($\phi$に)\end{CJK*}, such as canvas, papers and something to draw on.
Actually these predicate and argument preferences are numerous.
If PAS analysis models learn these preferences, they become strong clues for detecting latent arguments in sentences.
However, it is very difficult for PAS analysis models to learn these numerous relationships from limited size of annotated corpus.

\end{comment}

\section{Task Description}

%\subsection{Japanese Anaphora Resolution}
\begin{figure*}[t]
	%\hspace{-3em}
	\includegraphics[scale=0.9,clip]{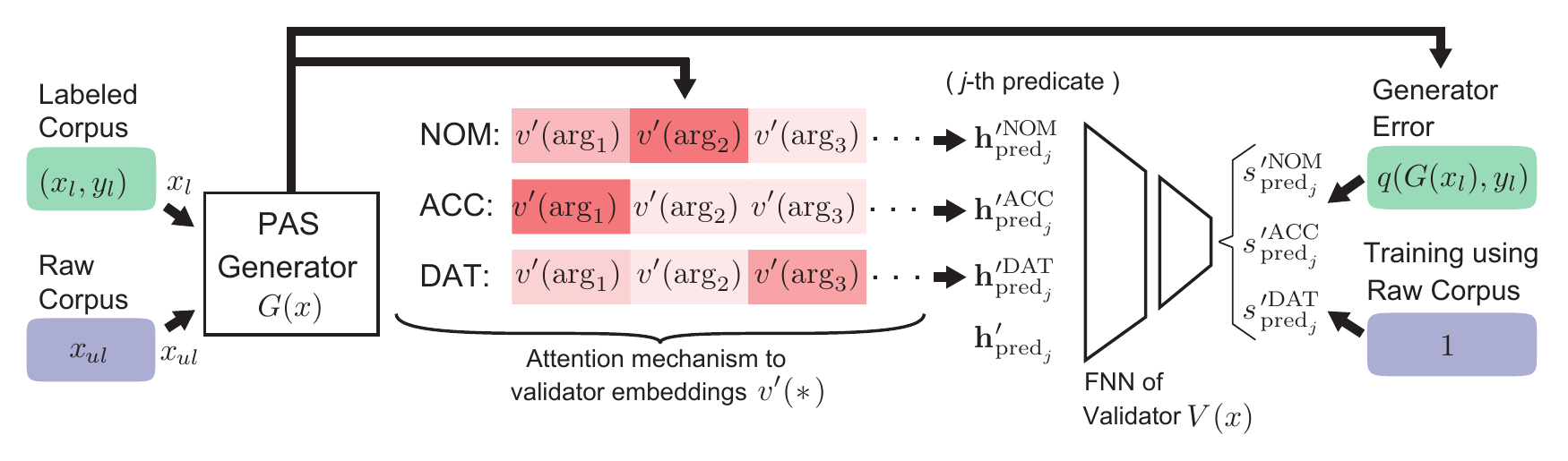}
       \vspace{-2.5em}
       \caption{
       The overall model of adversarial training with a raw corpus.
       The PAS generator $G(x)$ and validator $V(x)$.
       The validator takes inputs from the generator as a form of the attention mechanism.
       The validator itself is a simple feed-forward network with inputs of $j$-th predicate and its argument representations: 
       $\{h_{\mathrm{pred}_j}^{\prime}, h_{\mathrm{pred}_j}^{\prime \mathrm{case}_k}\}$.
       The validator returns scores for three cases and they are used for both the supervised training of the validator and the unsupervised training of the generator.
       The supervised training of the generator is not included in this figure.
       }
       \label{fig_overall}
\end{figure*}

%We focus on Japanese PAS analysis based on neural netwrok models.
Japanese PAS analysis determines essential case roles of words for each predicate:
\textit{who} did \textit{what} to \textit{whom}.
In many languages, such as English,
case roles are mainly determined by word order.
However, in Japanese, word order is highly flexible.
In Japanese, major case roles are the nominative case (NOM),
the accusative case (ACC) and the dative case (DAT), which
%These cases are
roughly correspond to Japanese surface case markers:
{\small \begin{CJK*}{UTF8}{ipxm}\underline{が}\end{CJK*}(ga)},
{\small \begin{CJK*}{UTF8}{ipxm}\underline{を}\end{CJK*}(wo)}, and
{\small \begin{CJK*}{UTF8}{ipxm}\underline{に}\end{CJK*}(ni)}.
%Especially, in Japanese, PAS analysis
%However, Japanese PAS analysis is known as a difficult task because surface case markers 
%However, 
    These case markers are often hidden by topic markers, and case arguments are also often omitted.
%\cite{ouchi2015,shibata2016}.
% relative clauses

We explain two detailed tasks of PAS analysis: case analysis and zero anaphora resolution.
In Table \ref{table:examples}, we show four example Japanese sentences and their PAS labels.
PAS labels are attached to nominative, accusative and dative cases of each predicate.
Sentence
(1) 
%{ \small``\begin{CJK*}{UTF8}{ipxm}タクシー\underline{が}(takushi-ga)~客\underline{を}(kyaku-wo)~駅\underline{に}(eki-ni)~送った(okutta)~。\end{CJK*}''}
has surface case markers that correspond to argument cases.

Sentence (2) is an example sentence for case analysis.
Case analysis is a task to find hidden case markers of arguments that have direct dependencies to their predicates.
Sentence (2) 
%{\small``\begin{CJK*}{UTF8}{ipxm}その~列車\underline{は}~荷物\underline{を}$_{\mathrm{ACC}}$~運んだ~。\end{CJK*}''}   \\
does not have the nominative case marker {\small \begin{CJK*}{UTF8}{ipxm}\underline{が}\end{CJK*}(ga)}.
It is hidden by the topic case marker
{\small \begin{CJK*}{UTF8}{ipxm}\underline{は}\end{CJK*}(wa)}.
Therefore, a case analysis model has to find the correct NOM case argument
{\small \begin{CJK*}{UTF8}{ipxm}列車\end{CJK*}(train)}.

%The sentence (2) have a nominative argument as we explained in Table \ref{table:examples}.
%and

Sentence (3) is an example sentence for zero anaphora resolution.
%(3) {\small``\begin{CJK*}{UTF8}{ipxm}タクシー\underline{が}$_{\mathrm{NOM}}$~客\underline{を}$_{\mathrm{ACC}}$~乗せた~とき~事故\underline{に}$_{\mathrm{DAT}}$~巻き込まれた~。\end{CJK*}''} \\
Zero anaphora resolution is a task to find arguments that do not have direct dependencies to their predicates.
At the second predicate
{\small``\begin{CJK*}{UTF8}{ipxm}巻き込まれた\end{CJK*}''(was involved)},
the correct nominative argument is
{\small``\begin{CJK*}{UTF8}{ipxm}タクシー\end{CJK*}''(taxi)}, while this does not have direct dependencies to the second predicate.
A zero anaphora resolution model has to find {\small``\begin{CJK*}{UTF8}{ipxm}タクシー\end{CJK*}''(taxi)} from the sentence, and
assign it to the NOM case of the second predicate.

%It also sometimes happens that predicates and their case arguments are quite far away in natural sentences.
%Those become harder problems.
%}

%Following \newcite{hangyo2013}, 
In the zero anaphora resolution task,
some correct arguments are 
% exist in preceding sentences, or
%It also happens that dropped arguments exist, but 
not specified in the article.
This is called as \textit{exophora}.
We consider ``author'' and ``reader'' arguments as exophora \cite{hangyo2013}. 
%We add
%special representations of exophora, an anaphora that does not exist in the sentence, are used.
%Specifically, we add writer and read expressions.
They are frequently dropped from Japanese natural sentences.
%The last sentence
Sentence (4)  is an example of dropped nominative arguments.
%(4) {\small``\begin{CJK*}{UTF8}{ipxm}この(kono)~列車(ressha)~\underline{に}(ni)~\underline{は}(wa)~乗れません(noremasen)~。\end{CJK*}''}.
%The former sentence has a drop of nominative arguments of the first predicate {\small``\begin{CJK*}{UTF8}{ipxm}乗せた\end{CJK*}''} (carrying).
%Actually, the second predicate {\small``\begin{CJK*}{UTF8}{ipxm}巻き込まれた\end{CJK*}''} (involved) has the same nominative case with modified case marker.
%The latter sentence is an example of dropped nominative case.
In this sentence, the nominative argument is {\small``\begin{CJK*}{UTF8}{ipxm}あなた\end{CJK*}''} (you), but
{\small``\begin{CJK*}{UTF8}{ipxm}あなた\end{CJK*}''} (you) does not appear in the sentence.
This is also included in zero anaphora resolution.
Except these special arguments of exophora, we focus on intra-sentential anaphora resolution in the same way as \cite{shibata2016,iida2016,ouchi2017,matsubayashi2017}.
We also attach {\usefont{T1}{pcr}{m}{n} NULL} labels to cases that predicates do not have.
%Precisely speaking, the {\usefont{T1}{pcr}{m}{n} NULL}  case of a predicate can be divided into two types.
%The first type is that the predicate normally requires that case but it does not appear in the sentence.
%The second type is that the predicate does not have the case.

% (3) {\small``\begin{CJK*}{UTF8}{ipxm}客を~乗せた~タクシーも~事故に~巻き込まれた~。\end{CJK*}''}

%About : JAR
%Ga-case, Wo-case, Ni-case
%Japanese language has several case markers: 
%\begin{CJK*}{UTF8}{ipxm}が\end{CJK*}(ga),
%\begin{CJK*}{UTF8}{ipxm}を\end{CJK*}(wo),
%\begin{CJK*}{UTF8}{ipxm}に\end{CJK*}(ni).
%Each predicate has nominative-case (ga-case), accusative case (wo-case), and dative case (ni-case).
%In naive viewpoint, cases are relative to certain case markers 

\section{Model}

%\subsection{Head Selection-based model}
\subsection{Generative Adversarial Networks}

Generative adversarial networks are originally proposed in image generation tasks \cite{goodfellow2014,salimans2016,springenberg2015}.
In the original model in \newcite{goodfellow2014}, they propose a generator $G$ and a discriminator $D$.
The discriminator $D$ is trained to devide the real data distribution $p_{data}(\mathbf{x})$ and
images generated from the noise samples $\mathbf{z}^{(i)} \in \mathcal{D}_{\mathbf{z}}$ from noise prior $p(\mathbf{z})$.
The discriminator loss is
\begin{align}
    \mathcal{L}_{D}=-\bigl( \mathbb{E}_{\mathbf{x} \sim p_{data}(\mathbf{x})}[\log D(\mathbf{x})] \nonumber \\
    + \mathbb{E}_{\mathbf{z} \sim p_z(\mathbf{z})}[\log(1- D(G(\mathbf{z})))] \bigl)~~,
\end{align}
and they train the discriminator by minimizing this loss while fixing the generator $G$.
Similarly, the generator $G$ is trained through minimizing
\begin{align}
    \mathcal{L}_{G}=\frac{1}{|\mathcal{D}_{\mathbf{z}}|} \sum_{i} \Bigl[ \log\left(1-D(G(\mathbf{z}^{(i)}))\right) \Bigl]~~,
\end{align}
    %\mathcal{L}_{G}=\mathbb{E}_{\mathbf{z} \sim p_z(\mathbf{z})}[\log\left(1-D(G(\mathbf{z}))\right)]~~,
while fixing the discriminator $D$.
By doing this, the discriminator tries to descriminate the generated images from real images, while 
the generator tries to generate images that can deceive the adversarial discriminator.
%\min_G \max_D D(G,D)=E_{x \sim p_{data}(\mathbf{x})}[\log D(\mathbf{x}))] + ...
This training scheme is applied for many generative tasks including
sentence generation \cite{subramanian2017},
machine translation \cite{britz-le-pryzant:2017:WMT},
dialog generation \cite{li-EtAl2017},
and text classification \cite{liuqiuhuang2017}.
%In such work,
%generator

%In many previous GAN research, the discriminators are training against the labels whether the input data is real data $\mathbf{x}$ or the generated distribution $G(p(\mathbf{z}))$ \cite{springenberg2015}.
%This method is effective especially in image generation, sentence generation or paraphrase generation.

\subsection{Proposed Adversarial Training Using Raw Corpus}
Japanese PAS analysis and many other syntactic analyses in NLP
are not purely generative, and
we can make use of a raw corpus instead of the numerical noise distribution $p(\mathbf{z})$.
In this work, we use an adversarial training method using a raw corpus, combined with ordinary supervised learning using an annotated corpus.
Let $\mathbf{x}_l \in \mathcal{D}_{l}$ indicate labeled data and $p(\mathbf{x}_l)$ indicate their label distribution.
We also use unlabeled data $\mathbf{x}_{ul} \in \mathcal{D}_{ul} $ later.
Our generator $G$ can be trained by the cross entropy loss with labeled data:
\begin{align}
    \mathcal{L}_{G/SL}=-\mathbb{E}_{ \mathbf{x}_l, y \sim p(\mathbf{x}_{l})} \bigl[\log G(\mathbf{x}_l) \bigl]~~.
\label{eq:cc}
\end{align}
Supervised training of the generator works by minimizing this loss.
Note that we follow the notations of \newcite{subramanian2017} in this subsection.

In addition, we train a so-called \textit{validator} 
against the generator errors.
We use the term ``validator'' instead of ``discriminator'' for our adversarial training.
Unlike the discriminator that is used for dividing generated images and real images, our validator is used to score the generator results.
Assume that $\mathbf{y}_l$ is the true labels and $G(\mathbf{x}_l)$ is the predicted label distribution of data $\mathbf{x}_l$ from the generator.
We define the labels of the generator errors as:
\begin{align}
    q(G(\mathbf{x}_l),\mathbf{y}_l)=\delta_{\argmax[G(\mathbf{x}_l)],~\mathbf{y}_l}~~,
\label{eq:error}
\end{align}
    %\mathbf{q}(G(\mathbf{x}_l),\mathbf{y}_l)={\bm \delta}_{\argmax[G(\mathbf{x}_l)],~\mathbf{y}_l}~~,
where $\delta_{i,j}=1$ only if $i=j$, otherwise $\delta_{i,j}=0$.
%Note that $1$ indicates that the generator prediction is correct and $0$ indicates that the generator prediction is incorrect.
This means that $q$ is equal to 1 if the argument that the generator predicts is correct, otherwise 0.
We use this generator error for training labels of the following validator.
%The adversarial training is 
% L=-E_{x \sim h(\mathbf{x}_{l})}[G(x)] + \\
The inputs of the validator are both the generator outputs $G(\mathbf{x})$ and data $\mathbf{x} \in \mathcal{D}$.
The validator can be written as $V(G(\mathbf{x}))$.
The validator $V$ is trained with labeled data $\mathbf{x}_l$ by
\begin{align}
    \mathcal{L}_{V/SL}=-\mathbb{E}_{\mathbf{x}_l, y \sim q(G(\mathbf{x}_l),\mathbf{y}_l)} \bigl[\log V(G(\mathbf{x}_l))\bigl]~~,
\label{eqn:disc}
%L_{V/SL}=-E_{G(\textbf{x}_l) \sim e(G(\mathbf{x}_l),\mathbf{y}_l)}[\log V(G(\mathbf{x}_l), \mathbf{x}_l)]~~,
\end{align}
while fixing the generator $G$.
This equation means that the validator is trained with labels of the generator error $q(G(\mathbf{x}_l),\mathbf{y}_l)$.
%that is also computed during the neural computation graph.

Once the validator is trained, we train the generator with an unsupervised method.
The generator $G$ is trained with unlabeled data $\mathbf{x}_{ul} \in \mathcal{D}_{ul}$ by minimizing the loss
\begin{align}
    \mathcal{L}_{G/UL}=- \frac{1}{|\mathcal{D}_{ul}|}\sum_{i} \bigl[\log V(G(\mathbf{x}_{ul}^{(i)}))\bigl]~~,
\label{eqn:raw}
\end{align}
%L_{G/UL}=-E_{\mathbf{x}_{ul}, y \sim 1}[\log (1- V(G(\mathbf{x}_{ul})))]~~,
%L_{G/UL}=-E_{G(\textbf{x}_{ul})\sim 1}[\log (1- V(G(\mathbf{x}_{ul}),\mathbf{x}_{ul}))]~~,
%L_{G}=E_{\mathbf{x} \sim \mathbf{x}_{ul}}[e(x_l,y_l)-V(G[x])]
while fixing the validator $V$.
This generator training loss using the validator can be explained as follows.
%the generator is trained by 
The generator tries to increase the validator scores to 1, while the validator is fixed.
If the validator is well-trained, it returns scores close to 1 for correct PAS labels that the generator outputs,
and 0 for wrong labels.
Therefore, in Equation (\ref{eqn:raw}), the generator tries to predict correct labels in order to increase the scores of fixed validator.
Note that the validator has a sigmoid function for the output of scores.
Therefore output scores of the validator are in $\left[0,1\right]$.

%The first term is the training of $G$ using raw data,
%while the second term is the training of $V$ using the output of $G$ and the Eq. \ref{eq:error}.

%When we train the

We first conduct the supervised training of generator network with Equation (\ref{eq:cc}).
After this,
following \newcite{goodfellow2014}, 
we use $k$-steps of the validator training and one-step of the generator training.
We also alternately conduct $l$-steps of supervised training of the generator.
The entire loss function of this adversarial training is
\begin{align}
\mathcal{L}=\mathcal{L}_{G/SL}+\mathcal{L}_{V/SL}+\mathcal{L}_{G/UL}~~.
\end{align}
%\textbf{
%Adversarial training using raw corpus works by minimising this loss.
%}

Our contribution is that we propose the validator and train it against the generator errors,
instead of discriminating generated data from real data.
\newcite{salimans2016} explore the semi-supervised learning using adversarial training for $K$-classes image classification tasks.
They add a new class of images that are generated  by the generator and classify them.

% combination of supervised training and generative learning.
\newcite{miyato2016} propose virtual adversarial training for semi-supervised learning.
They exploit unlabeled data for continuous smoothing of data distributions based on the adversarial perturbation of \newcite{goodfellow2015}.
These studies, however, do not use the counterpart neural networks for learning structures of unlabeled data.

%In Section 
%\ref{sec:related},
%we compare our model with reinforcement learning models.

In our Japanese PAS analysis model,
%Our neural network model consists of two parts:
the generator corresponds to the head-selection-based neural network for Japanese anaphora resolution.
%and the validator model.
Figure \ref{fig_overall} shows the entire model.
The labeled data correspond to the annotated corpora and the labels correspond to the PAS argument labels.
The unlabeled data correspond to raw corpora.
We explain the details of the generator and the validator neural networks in Sec.\ref{sec:gen} and Sec.\ref{sec:val} in turn.

\subsection{Generator of PAS Analysis}
\label{sec:gen}

\begin{figure}[t]
	\hspace{-1.0em}
	\includegraphics[scale=0.9,clip]{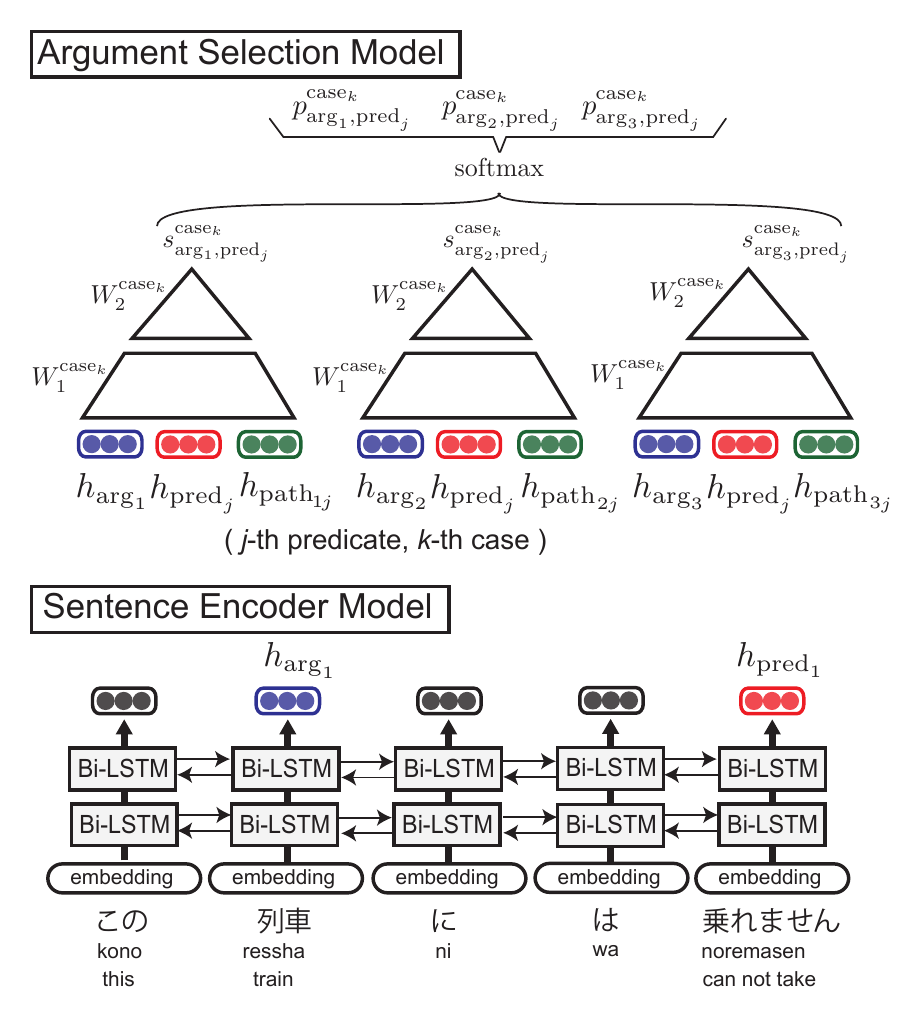}
	\vspace{-2.0em}
       \caption{
           The generator of PAS. 
           The sentence encoder is a three-layer bi-LSTM to compute the distributed representations of a predicate and its arguments: $h_{\mathrm{pred}_i}$ and $h_{\mathrm{arg}_i}$.
           The argument selection model is two-layer feedforward neural networks to compute the scores, $s_{\mathrm{arg}_i,\mathrm{pred}_j}^{\mathrm{case}_k}$, of candidate arguments for each case of a predicate.
       }
       \label{fig:gen1}
    %\vspace{-1em}
\end{figure}

%Japanese Anaphora resolutionを予測するニューラルネットワークは、3層のBi-LSTMと、FNN with softmaxからなる。
%The generator model for PAS analysis is rather complex neural networks that uses word and POS tag embeddings and syntactic information.

The generator predicts the probabilities of arguments for each of the NOM, ACC and DAT cases of a predicate.
%The generator is a neural network model that is based on sentence-wide RNNs and the head-selection model for
%choosing predicate and argument pairs.
As shown in Figure \ref{fig:gen1}, the generator consists of a sentence encoder and an argument selection model.
 %(a) a three-layer bi-LSTM and (b) the following two layer feedforward neural network and softmax function.
In the sentence encoder, we use a three-layer bidirectional-LSTM (bi-LSTM) to read the whole sentence and extract both global and local features as distributed representations.
The argument selection model consists of a two-layer feedforward neural network (FNN) and a softmax function.

For the sentence encoder, inputs are given as a sequence of embeddings $v(x)$, each of which consist of word $x$, its inflection from, POS and detailed POS.
They are concatenated and fed into the bi-LSTM layers.
The bi-LSTM layers read these embeddings in forward and backward order
and outputs the distributed representations of a predicate and a candidate argument: $h_{\mathrm{pred}_j}$ and $h_{\mathrm{arg}_i}$.
Note that we also use the exophora entities, i.e., an author and a reader, as argument candidates.
Therefore, we use specific embeddings for them.
These embeddings are not generated by the bi-LSTM layers but 
are directly used in the argument selection model.

We also use path embeddings to capture a dependency relation between a predicate and its candidate argument as used in \newcite{roth2016}.
%We also compute path embeddings of predicates and candidate arguments with a bi-LSTM layer by the following procedure.
Although \newcite{roth2016} use a one-way LSTM layer to represent the dependency path from a predicate to its potential argument,
we use a bi-LSTM layer for this purpose.
We feed the embeddings of words and POS tags to the bi-LSTM layer.
In this way, the resulting path embedding represents both predicate-to-argument and argument-to-predicate paths.
% The bi-LSTM layer for path embeddings reads the dependency path and outputs the representations of the predicate $j$ and its candidate argument $i$.
We concatenate the bidirectional path embeddings to generate $h_{\mathrm{path}_{ij}}$, which represents the dependency relation between the predicate $j$ and its candidate argument $i$.

For the argument selection model,
we apply the argument selection model \cite{zhang-cheng-lapata2017} to evaluate the relation between a predicate and its potential argument for each argument case.
In the argument selection model, a single FNN is repeatedly used to calculate scores for a child word and its head candidate word,
and then a softmax function calculates normalized probabilities of candidate heads.
We use three different FNNs that correspond to the
NOM, ACC and DAT cases.
These three FNNs have the same inputs of the distributed representations of 
$j$-th predicate $h_{\mathrm{pred}_j}$, $i$-th candidate argument $h_{\mathrm{arg}_i}$ and path embedding $h_{\mathrm{path}_{ij}}$ between the predicate $j$ and candidate argument $i$.
The FNNs for NOM, ACC and DAT compute the argument scores $s_{\mathrm{arg}_i,\mathrm{pred}_j}^{\mathrm{case}_k}$,
where $\mathrm{case}_k \in \{\mathrm{NOM}, \mathrm{ACC}, \mathrm{DAT}\}$.
Finally, the softmax function computes the probability
$p({\scriptstyle\mathrm{arg}_i}|{\scriptstyle\mathrm{pred}_j,\mathrm{case}_k})$
of candidate argument $i$ for case $k$ of $j$-th predicate
as:
\begin{align}
	p({\scriptstyle\mathrm{arg}_i}|{\scriptstyle\mathrm{pred}_j,\mathrm{case}_k}) =
 \frac{\exp \left(s_{\mathrm{arg}_i,\mathrm{pred}_j}^{\mathrm{case}_k} \right)}
{\displaystyle \sum_{\mathrm{arg}_i} \exp \left(s_{\mathrm{arg}_i,\mathrm{pred}_j}^{\mathrm{case}_k} \right)} .
\label{eqn:gen}
\end{align}

Our argument selection model is similar to the neural network structure of \newcite{matsubayashi2017}.
However, \newcite{matsubayashi2017} does not use RNNs to read the whole sentence.
Their model is also designed to choose a case label for a pair of a predicate and its argument candidate.
In other words, their model can assign the same case label to multiple arguments by itself, while our model does not.
Since case arguments are almost unique for each case of a predicate in Japanese,
\newcite{matsubayashi2017} select the argument that has the highest probability for each case, even though probabilities of case arguments are not normalized over argument candidates.
The model of \newcite{ouchi2017} has the same problem.

%Note that we add special representations from the outer world:

\subsection{Validator}
\label{sec:val}

We exploit a validator to train the generator using a raw corpus.
%The validator is a rather simple neural network compared to the generator.
It consists of a two-layer FNN to which
embeddings of a predicate and its arguments are fed.
For predicate $j$, the input of the FNN is the representations of the predicate $h_{\mathrm{pred}_j}^{\prime}$ and three arguments
%$\left\{h_{\mathrm{arg}}^{\prime~\mathrm{DAT}}, h_{\mathrm{arg}}^{\prime~\mathrm{ACC}}, h_{\mathrm{\mathrm{arg}}}^{\prime~\mathrm{NOM}}\right\}$.
$\left\{h_{\mathrm{pred}_j}^{\prime~\mathrm{NOM}}, h_{\mathrm{pred}_j}^{\prime~\mathrm{ACC}}, h_{\mathrm{pred}_j}^{\prime~\mathrm{DAT}}\right\}$
that are inferred by the generator.
The two-layer FNN outputs three values, and then
three sigmoid functions compute the scores of scalar values in a range of $\left[0,1\right]$ for the NOM, ACC and DAT cases:
$\left\{s_{\mathrm{pred}_j}^{\prime~\mathrm{NOM}}, s_{\mathrm{pred}_j}^{\prime~\mathrm{ACC}}, s_{\mathrm{pred}_j}^{\prime~\mathrm{DAT}}\right\}$.
%Each score of the validator is a scalar value and  because of the last sigmoid function.
These scores are the outputs of the validator $D(x)$.
We use dropout of 0.5 at the FNN input and hidden layer.
% \footnote{
% The number of validator outputs is the number of cases times the number of predicates in the sentence.}
% \textbf{For each predicate in the sentence, we adapt this and output scores.}

The generator and validator networks are coupled by the attention mechanism, or the weighted sum of the validator embeddings.
% \cite{bahdanau2014}.
As shown in Equation (\ref{eqn:gen}), we compute a probability distribution of candidate arguments.
We use the weighted sum of embeddings $v'(x)$ of candidate arguments to compute the input representations of the validator:
% $h_{\mathrm{arg}_i}^{\prime~\mathrm{case}_k}$:
\begin{align}
	h_{\mathrm{pred}_j}^{\prime~\mathrm{case}_k} &= E_{\mathbf{x} \sim p(\mathrm{arg}_i)}[v'(\mathbf{x})] \nonumber \\
 &= \sum_{{\scriptstyle\mathrm{arg}_i}} p({\scriptstyle\mathrm{arg}_i}|{\scriptstyle\mathrm{pred}_j,\mathrm{case}_k})v'({\scriptstyle\mathrm{arg}_i}) . \nonumber
\end{align}
This summation is taken over candidate arguments in the sentence and the exophora entities.
Note that we use embeddings $v'(x)$ for the validator that are different from the embeddings $v(x)$ for the generator,
in order to separate the computation graphs of the generator and the validator neural networks except the joint part.
We use this weighted sum by the softmax outputs instead of the argmax function.
This allows the backpropagation through this joint.
We also feed the embedding of a predicate to the validator:
\begin{align}
	h_{\mathrm{pred}_j}^{\prime} &= v'({\scriptstyle\mathrm{pred}_j}).
\end{align}

Note that the validator is a simple neural network compared with the generator.
The validator has limited inputs of predicates and arguments and no inputs of other words in sentences.
This allows the generator to overwhelm the validator during the adversarial training.

% \textbf{This is given from the annotated or automatically parsed tags of the corpus, and
% is not affected by the generator outputs.}

%hogehoge
%\begin{align}
% \sum_{\mathrm{arg}_i} [p(\mathrm{arg}_i|\mathrm{pred}_j,\mathrm{case}_k)v() ]
%\end{align}

%This validator could learn possible combinations of predicates and arguments.

\subsection{Implementation Details}
\label{sec:imple}
The neural networks are trained using backpropagation.
The backpropagation has been done to the word and POS tags.
We use Adam \cite{kingma2015adam} at the initial training of the generator network for the gradient learning rule.
In adversarial learning,
Adagrad \cite{duchi2010adagrad} is suitable because of the stability of learning.
We use pre-trained word embeddings from 100M sentences from Japanese web corpus by word2vec \cite{mikolov2013}.
Other embeddings and hidden weights of neural networks are randomly initialized.

\begin{table}[tb]
\begin{center}
	\footnotesize\begin{tabular}{lc}
            \toprule[1.5pt]
    Type                             & Value \\ \midrule
    Size of hidden layers of FNNs     & 1,000 \\
    Size of Bi-LSTMs               & 256 \\
    Dim. of word embedding      & 100 \\
    Dim. of POS, detailed POS, inflection form tags     & 10,~10,~9 \\
    Minibatch size for the generator and validator      & 16,~1 \\
            \bottomrule[1.5pt]
	\end{tabular}
    \caption{
    Parameters for neural network structure and training.
    }
    \label{table:params}
    \end{center}
\end{table}
    %Embedding vocabulary size      &  \\
    %Training steps of the generator $k$      & 16 \\
    %Training steps of the validator $l$      & 4 \\
 
For adversarial training, we first train the generator for two epochs by the supervised method,
and train the validator while fixing the generator for another epoch.
This is because the validator training preceding the generator training makes the validator result worse.
After this, we alternately do the unsupervised training of the generator ($L_{G/UL}$),
$k$-times of supervised training of the validator ($L_{V/SL}$) and
$l$-times of supervised training of the generator ($L_{G/SL}$).

We use the
$N(L_{G/UL})/N(L_{G/SL})=1/4$
and
$N(L_{V/SL})/N(L_{G/SL})=1/4$,
where $N(\cdot)$ indicates the number of sentences used for training.
Also we use minibatch of 16 sentences for both supervised and unsupervised training of the generator,
while we do not use minibatch for validator training.
Therefore, we use $k=16$ and $l=4$.
%This means that one sentence of unsupervised training of the generator and one sentence of 
%for four sentences of supervised 
%Therefore one sentence of unsupervised training of the generator and one supervised training of the validator is taken place,
%during four sentences of supervised training of the generator is taken place (NEED TO BE REVISED).
Other parameters are summarized in Table \ref{table:params}.

\section{Experiments}

\begin{table}[t]
\begin{center}
	\begin{tabular}{cccc}
        \toprule[1.5pt]
        KWDLC & \# snt & \# of dep & \# of zero \\
        \midrule
        Train & 11,558 & 9,227 & 8,216 \\
        Dev.  & 1,585  & 1,253  & 821 \\
        Test  & 2,195  & 1,780  & 1,669 \\
        \bottomrule[1.5pt]
	\end{tabular}
    \\
    \caption{KWDLC data statistics.}
    \vspace{-1em}
    \label{table:stat}
\end{center}
\end{table}

\begin{table}[t]
\begin{center}
	\begin{tabular}{cccc}
        \toprule[1.5pt]
        KWDLC & NOM & ACC & DAT \\
        \midrule
        \# of dep  & 7,224 & 1,555 & 448 \\
        \# of zero & 6,453 & 515  & 1,248 \\
        \bottomrule[1.5pt]
	\end{tabular}
    \\
    \caption{KWDLC training data statistics for each case.
    %We count arguments that are used for training.
	}
    \label{table:case}
    \vspace{-1em}
\end{center}
\end{table}

\begin{table}[t]
\begin{center}
	\begin{tabular}{lcc}
        \toprule[1.5pt]
        & Case &  Zero \\
        \midrule
        Ouchi+ 2015    & 76.5 & 42.1 \\
        Shibata+ 2016  & 89.3 & 53.4 \\
        \midrule
        Gen            & 91.5 & 56.2 \\
        Gen+Adv        & \textbf{92.0}$^\ddagger$ & \textbf{58.4}$^\ddagger$ \\
        \bottomrule[1.5pt]
	\end{tabular}
    \caption{
    The results of case analysis (Case) and zero anaphora resolution (Zero).
    %All values are evaluated in F-measure.
    We use F-measure as an evaluation measure.
    $\ddagger$ denotes that the improvement is statistically significant at $p<0.05$, compared with Gen using paired t-test.
    }
    \vspace{-1.5em}
    \label{table:resultall}
\end{center}
\end{table}

\begin{table*}[t]
\begin{center}
	\begin{tabular}{lcccccc}
        \toprule[1.5pt]
        & \multicolumn{3}{c}{Case analysis} &  \multicolumn{3}{c}{Zero anaphora resolution} \\
        \cmidrule(r{4pt}){2-4} \cmidrule(l){5-7}
        Model         & NOM  & ACC  & DAT  & NOM  & ACC  & DAT  \\
        \midrule                                                
        Ouchi+ 2015   & 87.4 & 40.2 & 27.6 & 48.8 &  0.0 & 10.7 \\
        Shibata+ 2016 & 94.1 & 75.6 & 30.0 & 57.7 & 17.3 & 37.8 \\
        Gen           & \textbf{95.3} & 83.6 & 39.7 & 60.7 & 30.4 & 41.2 \\
        Gen+Adv       & \textbf{95.3} & \textbf{85.4} & \textbf{51.5} & \textbf{62.3} & \textbf{31.1} & \textbf{44.6} \\
        \bottomrule[1.5pt]
	\end{tabular}
    \caption{
    The detailed results of case analysis and zero anaphora resolution for the NOM, ACC and DAT cases.
    Our models outperform the existing models in all cases.
    All values are evaluated with F-measure.
    }
    \vspace{-1em}
    \label{table:result}
\end{center}
\end{table*}

\begin{table}[t]
\begin{center}
	\begin{tabular}{lcc}
        \toprule[1.5pt]
        & Case &  Zero \\
        \midrule                                                
        Gen & 91.5 & 56.2 \\
        Gen+Aug & 91.2 & 57.0 \\
        \midrule                                                
        Gen+Adv        & \textbf{92.0}$^\ddagger$ & \textbf{58.4}$^\ddagger$ \\
        \bottomrule[1.5pt]
	\end{tabular}
    \caption{
        The comparisons of Gen+Adv with Gen and the data augmentation model (Gen+Aug).
    % case analysis (Case) and zero anaphora resolution (Zero).
    %All values are evaluated in F-measure.
    %We use F-measure as an evaluation measure.
    $\ddagger$ denotes that the improvement is statistically significant at $p<0.05$, compared with Gen+Aug.
    }
    \vspace{-0.5em}
    \label{table:resultaug}
\end{center}
\end{table}

\subsection{Experimental Settings}

Following \newcite{shibata2016}, we use the KWDLC (Kyoto University Web Document Leads Corpus) corpus \cite{hangyo2012} for our experiments.\footnote{
The KWDLC corpus is available at \url{http://nlp.ist.i.kyoto-u.ac.jp/EN/index.php?KWDLC}}
This corpus contains various Web documents, such as news articles, personal blogs, and commerce sites.
In KWDLC, lead three sentences of each document are annotated with PAS structures including zero pronouns.
For a raw corpus, we use a Japanese web corpus created by \newcite{hangyo2012}, which has no duplicated sentences with KWDLC.
%\footnote{This raw corpus will be published upon acceptance.}
This raw corpus is automatically parsed by the Japanese dependency parser KNP.

% KWDLC also includes other exophora, such as unspecified people.
We focus on intra-sentential anaphora resolution, and so we apply a preprocess to KWDLC.
%In this preprocess, 
We regard the anaphors whose antecedents are in the preceding sentences as {\usefont{T1}{pcr}{m}{n} NULL} in the same way as \newcite{ouchi2015,shibata2016}.
Tables \ref{table:stat} and \ref{table:case} list the statistics of KWDLC.
% \textbf{Note that we count arguments that are used for training.}

We use the exophora entities, i.e., an author and a reader, following the annotations in KWDLC.
We also assign author/reader labels to the following expressions in the same way as \newcite{hangyo2013,shibata2016}:
\begin{description}
\item[author] {\small ``\begin{CJK*}{UTF8}{ipxm}私\end{CJK*}''} (I), {\small ``\begin{CJK*}{UTF8}{ipxm}僕\end{CJK*}''} (I), {\small ``\begin{CJK*}{UTF8}{ipxm}我々\end{CJK*}''} (we), {\small ``\begin{CJK*}{UTF8}{ipxm}弊社\end{CJK*}''} (our company)
\item[reader] {\small ``\begin{CJK*}{UTF8}{ipxm}あなた\end{CJK*}''} (you), {\small ``\begin{CJK*}{UTF8}{ipxm}君\end{CJK*}''} (you), {\small ``\begin{CJK*}{UTF8}{ipxm}客\end{CJK*}''} (customer), {\small ``\begin{CJK*}{UTF8}{ipxm}皆様\end{CJK*}''} (you all)
\end{description}

%We also add a {\usefont{T1}{pcr}{m}{n} NULL}  for cases that predicates do not have.
%Precisely speaking, the {\usefont{T1}{pcr}{m}{n} NULL}  case of a predicate can be divided into two types.
%The first type is that the predicate normally requires that case but it does not appear in the sentence.
%The second type is that the predicate does not have the case.

Following \newcite{ouchi2015} and \newcite{shibata2016},
we conduct two kinds of analysis:
(1) case analysis and (2) zero anaphora resolution.
Case analysis is the task to determine the correct case labels when predicates and their arguments have direct dependencies
but their case markers are hidden by surface markers, such as topic markers.
Zero anaphora resolution is a task to find certain case arguments
that do not have direct dependencies to their predicates in the sentence.

Following \newcite{shibata2016}, we exclude predicates that the same arguments are filled in multiple cases of a predicate.
This is relatively uncommon and 1.5 \% of the whole corpus are excluded.
Predicates are marked in the gold dependency parses. Candidate arguments are just other tokens than predicates. This setting is also the same as \newcite{shibata2016}.
%These sentences are about 200 sentences (or 1.5 \%) in the whole data set.
%We use annotated morphological analysis and dependency labels.

All performances are evaluated with micro-averaged F-measure \cite{shibata2016}.

\subsection{Experimental Results}
We compare two models: the supervised generator model (Gen) and
the proposed semi-supervised model with adversarial training (Gen+Adv).
We also compare our models with two previous models: \newcite{ouchi2015} and \newcite{shibata2016}, whose performance on the KWDLC corpus is reported.

Table \ref{table:resultall} lists the experimental results.
Our models (Gen and Gen+Adv) outperformed the previous models.
Furthermore, the proposed model with adversarial training (Gen+Adv) was significantly better than the supervised model (Gen).

\subsection{Comparison with Data Augmentation Model}

We also compare our GAN-based approach with data augmentation techniques.
A data augmentation approach is used in \newcite{tingliu2017}.
They automatically process raw corpora and make drops of words with some rules.
However, it is difficult to directly apply their approach to Japanese PAS analysis because Japanese zero-pronoun depends on dependency trees.
If we make some drops of arguments of predicates in sentences, this can cause lacks of nodes in dependency trees.
If we prune some branches of dependency trees of the sentence, this cause the data bias problem.

Therefore we use existing training corpora and word embeddings for the data augmentation.
First we randomly choose an argument word $w$ in the training corpus and
then swap it with another word $w'$ with the probability of $p(w,w')$.
We choose top-20 nearest words to the original word $w$ in the pre-trained word embedding as candidates of swapped words.
The probability is defined as $ p(w,w') \propto [v(w)^\top v(w')]^r$, where $r=10$.
This probability is normalized by top-20 nearest words.
We then merge this pseudo data and the original training corpus and train the model in the same way with the Gen model.
We conducted several experiments and found that the model trained with the same amount of the pseudo data as the training corpus achieved the best result.

Table \ref{table:resultaug} shows the results of the data augmentation model and the GAN-based model.
Our Gen+Adv model performs better than the data augmented model.
Note that our data augmentation model does not use raw corpora directly.
%Therefore we cannot directly compare these results with that of \newcite{tingliu2017}.

\begin{figure*}[t]
	%\hspace{-3em}
\centering
%\subfloat[][]{
%   \rule{0.45\linewidth}{3cm}
% }
%\subfloat[][]{
%   \rule{0.45\linewidth}{3cm}
% }
    % \begin{minipage}[c]{\textwidth}
    %    \includegraphics[scale=0.3,clip]{graph/gen.pdf}
    %\end{minipage}
    % \begin{minipage}[c]{\textwidth}
    %    \includegraphics[scale=0.3,clip]{graph/disc.pdf}
    %\end{minipage}
  \begin{subfigure}[b]{0.4\textwidth}
      \hspace{-4em}
	\includegraphics[scale=0.70]{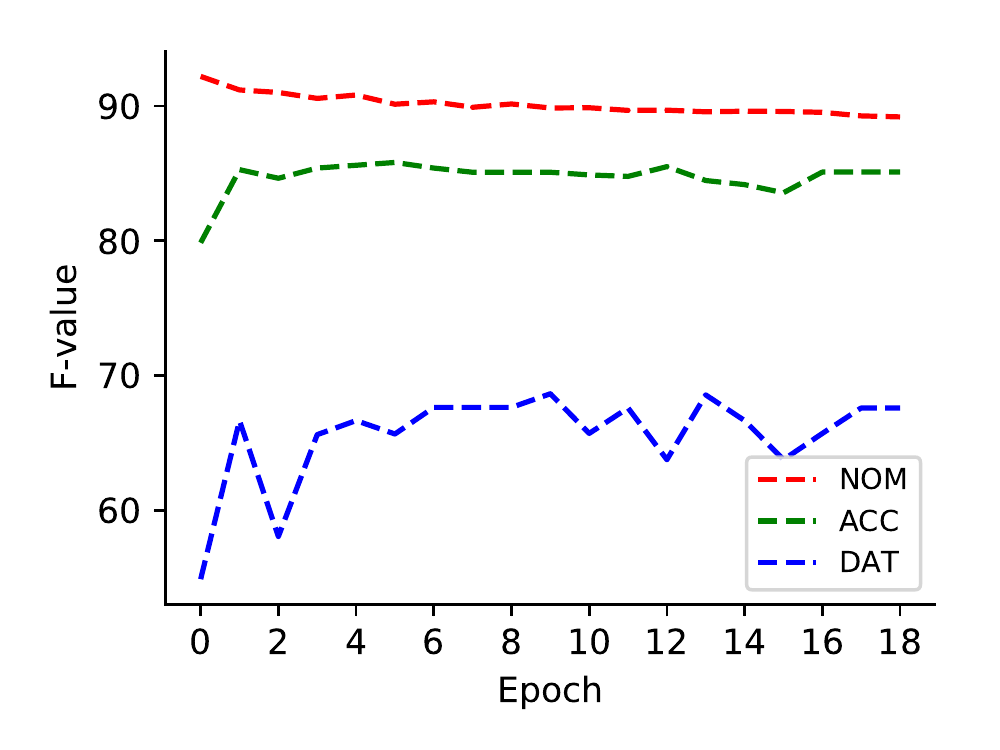}
    %\caption{Picture 1}
    %\label{fig:1}
  \end{subfigure}
  \begin{subfigure}[b]{0.4\textwidth}
      %\hspace{1em}
	\includegraphics[scale=0.70]{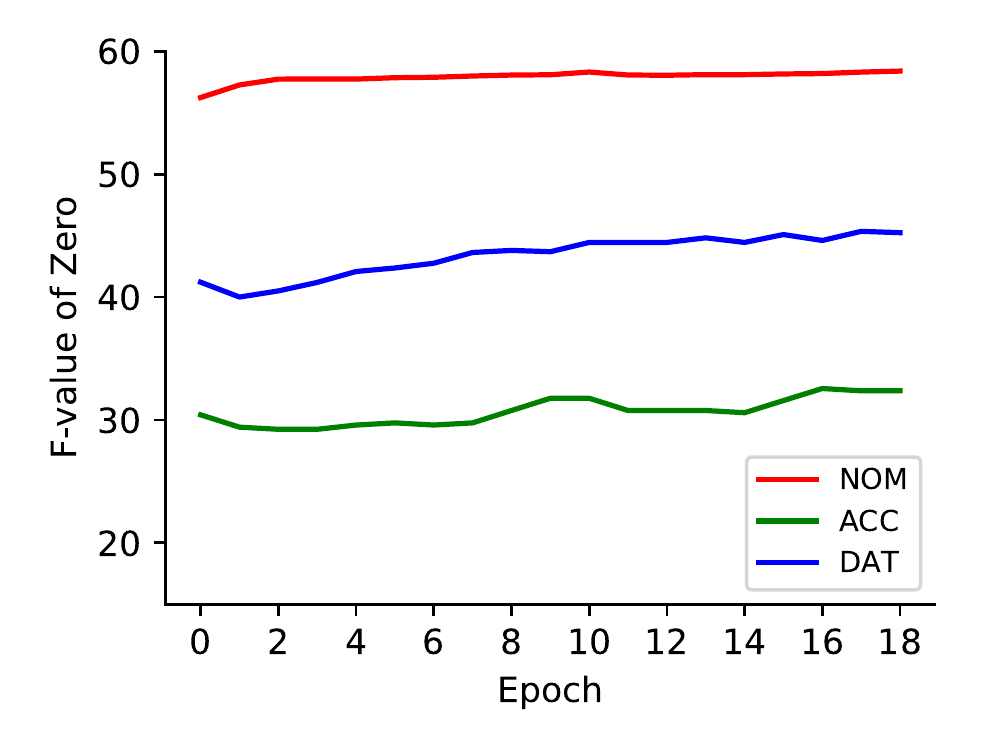}
    %\caption{Picture 2}
    %\label{fig:2}
  \end{subfigure}
       \caption{
    Left: validator scores with the development set during adversarial training epochs.
    Right: generator scores for Zero with the development set during adversarial training epochs.
       }
	\label{fig:disc}
\end{figure*}

\subsection{Discussion}
\subsubsection{Result Analysis}

We report the detailed performance for each case in Table \ref{table:result}.
Among the three cases, zero anaphora resolution of the ACC and DAT cases is notoriously difficult.
This is attributed to the fact that these ACC and DAT cases are fewer than the NOM case in the corpus as shown in Table \ref{table:case}.
However, we can see that our proposed model, Gen+Adv, performs much better than the previous models especially for the ACC and DAT cases.
%Actually, training data of ACC and DAT is much fewer than those of NOM .
%Therefore it is expected that our model can learn PAS analysis from very few examples.
Although the number of training instances of ACC and DAT is much smaller than that of NOM, our semi-supervised model can learn PAS for all three cases using a raw corpus.
This indicates that our model can work well in resource-poor cases.

We analyzed the results of Gen+Adv by comparing with Gen and the model of \newcite{shibata2016}.
Here, we focus on the ACC and DAT cases because
%they have much gains of scores.
their improvements are notable.

\begin{itemize}
    \item {\small ``\begin{CJK*}{UTF8}{ipxm}パック\underline{は}~~洗って、~~分別して~~リサイクルに~~出さなきゃいけないので~~手間がかかる。\end{CJK*}``} \\
                It is bothersome to wash, classify and recycle spent packs.
\end{itemize}
In this sentence, 
the predicates
{\small ``\begin{CJK*}{UTF8}{ipxm}洗って\end{CJK*}''}
(wash),
{\small ``\begin{CJK*}{UTF8}{ipxm}分別して\end{CJK*}''}
(classify),
{\small ``\begin{CJK*}{UTF8}{ipxm}(リサイクルに)~出す\end{CJK*}''}
(recycle)
takes the same ACC argument, 
{\small ``\begin{CJK*}{UTF8}{ipxm}パック\end{CJK*}''}
(pack).
This is not so easy for Japanese PAS analysis because the actual ACC case marker
{\small ``\begin{CJK*}{UTF8}{ipxm}\underline{を}\end{CJK*}''}
(wo) of
{\small ``\begin{CJK*}{UTF8}{ipxm}パック\end{CJK*}''} (pack)
is hidden by the topic marker
{\small ``\begin{CJK*}{UTF8}{ipxm}\underline{は}\end{CJK*}''}
(wa). The Gen+Adv model can detect the correct argument while 
the model of \newcite{shibata2016} fails.
In the Gen+Adv model, each predicate gives a high probability to 
{\small ``\begin{CJK*}{UTF8}{ipxm}パック\end{CJK*}''} (pack)
as an ACC argument and finally chooses this.
We found many examples similar to this and speculate that our model captures a kind of selectional preferences.
% Since these relations of predicates and preferred arguments are often seen in ACC and DAT cases,
% our model achieves better scores in these two cases.

The next example is an error of the DAT case by the Gen+Adv model.
\begin{itemize}
    \item {\small``\begin{CJK*}{UTF8}{ipxm}各専門分野も~~~お任せ下さい。\end{CJK*}''} \\
                please leave every professional field (to $\phi$)
\end{itemize}
%\begin{itemize}
%    \item DAT of gold: NULL
%    \item DAT of GEN : NULL
%    \item DAT of GEN+ADV: NULL
%\end{itemize}
% This is an error example from the test set.
The gold label of this DAT case (to $\phi$) is NULL because this argument is not written in the sentence.
% The Gen model also predicts it as NULL.
However, the Gen+Adv model judged the DAT argument as ``author''.
Although we cannot specify $\phi$ as ``author'' only from this sentence,
``author'' is a possible argument depending on the context.
%To solve this problem, we need to expand our model to inter-sentential PAS analysis.

\subsubsection{Validator Analysis}

We also evaluate the performance of the validator during the adversarial training with raw corpora.
Figure \ref{fig:disc} shows the validator performance and the generator performance of Zero on the development set.
The validator score is evaluated with the outputs of generator.
%This means that 
%Therefore, in a simple viewpoint, when the generator performs better,
%the validator performs worse because the validator is trained against the generator.

We notice that the NOM case and the other two cases have different curves in both graphs.
%In the validator scores, the curve of NOM decreases slightly, while ACC and DAT curves increase for the first few epochs,
%and then fluctuate.
%In the generator scores, the curve of NOM increases much in the first few epochs, and it slightly increases in later epochs.
%Other curves drop for the first few epochs and increase much in  the later epochs.
This can be explained by the speciality of the NOM case.
The NOM case has much more author/reader expressions than the other cases.
The prediction of author/reader expressions depends not only on selectional preferences of predicates and arguments but on the whole of sentences.
Therefore the validator that relies only on predicate and argument representations cannot predict author/reader expressions well.
%Japanese PAS analysis can be regarded as multi-task learning of case analysis and zero anaphora resolution.
%Therefore each case scores might affect each other.

In the ACC and DAT cases, the scores of the generator and validator increase in the first epochs.
This suggests that the validator learns the weakness of the generator and vice versa.
However, in later epochs, the scores of the generator increase with fluctuation, while the scores of the validator saturates.
%This suggest that, as the generator performs better, the validator also become better in the same extent of the generator.
%Otherwise, the scores of the validator get worse when the generator performs better.
This suggests that the generator gradually becomes stronger than the validator.

\section{Related Work}
\label{sec:related}

%\subsection{Japanese PAS Analysis}

%Paraphrase Generation with Deep Reinforcement Learning.

%Pseudo Large-scale data generation\cite{tingliu2017}.

%Recently, 
%PAS analysis in Japanese has gained plenty of attention in recent years.

\newcite{shibata2016} proposed a neural network-based PAS analysis model using local and global features.
This model is based on the non-neural model of \newcite{ouchi2015}.
They achieved state-of-the-art results on case analysis and zero anaphora resolution using the KWDLC corpus.
%In addition, 
They use an external resource to extract selectional preferences.
Since our model uses an external resource, we compare our model with the models of \newcite{shibata2016} and \newcite{ouchi2015}.

\newcite{ouchi2017} proposed a semantic role labeling-based PAS analysis model using Grid-RNNs.
\newcite{matsubayashi2017} proposed a case label selection model with feature-based neural networks.
%These models can assign the same label to multiple arguments, which is uncommon in natural Japanese sentences as discussed in Section 3.3.
They conducted their experiments on NAIST Text Corpus (NTC) \cite{iida2007,iida2016}.
NTC consists of newspaper articles, and does not include the annotations of author/reader expressions that are common in Japanese natural sentences.

\section{Conclusion}
We proposed a novel Japanese PAS analysis model that exploits a semi-supervised adversarial training.
The generator neural network learns Japanese PAS and selectional preferences,
while the validator is trained against the generator errors.
This validator enables the generator to be trained from raw corpora and enhance it with external knowledge.
In the future, we will apply this semi-supervised training method to other NLP tasks.
%, such as neural machine translation.
%We propose the joint parsing models
%by the feed-forward and bi-LSTM neural networks.
%Both of them use the character string embeddings.
%The character string embeddings help to capture the similarities of incomplete tokens.
%We also explore the neural network with few features using $n$-gram bi-LSTMs.
%Our SegTagDep joint model achieves better scores of Chinese word segmentation
%and POS tagging than previous joint models, and
%our SegTag and Dep pipeline model achieves state-of-the-art score of dependency parsing.
%The bi-LSTM models reduce the cost of feature engineering.

\section*{Acknowledgment}
This work was supported by JST CREST Grant Number JPMJCR1301, Japan and 
JST ACT-I Grant Number JPMJPR17U8, Japan.

\bibliography{kurita_acl2018}
\bibliographystyle{acl_natbib}

\appendix

%\section{Supplemental Material}
%\label{sec:supplemental}
%EACL-2017 also encourages the submission of supplementary material
%to report preprocessing decisions, model parameters, and other details
%necessary for the replication of the experiments reported in the 
%paper. Seemingly small preprocessing decisions can sometimes make
%a large difference in performance, so it is crucial to record such
%decisions to precisely characterise state-of-the-art methods.
%
%Nonetheless, supplementary material should be supplementary (rather
%than central) to the paper. It may include explanations or details
%of proofs or derivations that do not fit into the paper, lists of
%features or feature templates, sample inputs and outputs for a system,
%pseudo-code or source code, and data. (Source code and data should
%be separate uploads, rather than part of the paper).
%
%The paper should not rely on the supplementary material: while the paper
%may refer to and cite the supplementary material will be available to the
%reviewers, they will not be asked to review the
%supplementary material.
%
%Appendices (i.e. supplementary material in the form of proofs, tables,
%or pseudo-code) should come after the references, as shown here. Use
%\verb|\appendix| before any appendix section to switch the section
%numbering over to letters.
%
%\section{Multiple Appendices}
%\dots can be obtained by using more than one section. We hope you won't
%need that.

\end{document}